\def\year{2019}\relax
\documentclass[letterpaper]{article} 
\usepackage{aaai19}  
\usepackage{times}  
\usepackage{helvet}  
\usepackage{courier}  
\usepackage{url}  
\usepackage{graphicx}  
\usepackage{multirow}
\usepackage{makecell}
\usepackage{algorithm}
\usepackage{algorithmic}
\usepackage{color}
\usepackage{caption}
\usepackage{amsmath}
\usepackage{hyperref}
\usepackage{booktabs}
\usepackage{algorithm, algorithmic}
\begin{document}
%
\title{Structure Learning of Deep Networks via DNA Computing Algorithm }

\maketitle
\begin{abstract}
  Convolutional Neural Network (CNN) has gained state-of-the-art results in many pattern recognition and computer vision tasks. However, most of the CNN structures are manually designed by experienced researchers. Therefore, automatically building high performance networks becomes an important problem. In this paper, we introduce the idea of using DNA computing algorithm to automatically learn high-performance architectures. In DNA computing algorithm, we use short DNA strands to represent layers and long DNA strands to represent overall networks. We found that most of the learned models perform similarly, and only those performing worse during the first runs of training will perform worse finally than others. The indicates that: 1) Using DNA computing algorithm to learn deep architectures is feasible; 2) Local minima should not be a problem of deep networks; 3) We can use early stop to kill the models with the bad performance just after several runs of training. In our experiments, an accuracy 99.73\% was obtained on the MNIST data set and an accuracy 95.10\% was obtained on the CIFAR-10 data set.
\end{abstract}

\section{Introduction}
Convolutional neural network (CNN) has gained success in many fields and become the main method in pattern recognition and computer vision since AlexNet won the 2012 ImageNet competition \cite{krizhevsky2012imagenet,deng2009imagenet}. Nowadays, the technology has also been used in real-world applications, such as applications in smartphones, intelligent recommendations on products or music, target recognition and face detection \cite{lecun2015deep}.

However, designing a high performance architecture is still a challenging problem. This usually needs expertise and takes long time of research by trail and error \cite{hanxiao}. Because of numerous configurations with respect to the number of layers and details in each layer, it is impossible to explore the whole architecture space. Hence, how to automatically generate a competitive architecture becomes an important problem \cite{bergstra2012random,miller1989designing,negrinho2017deeparchitect}.

Reinforcement Learning and Evolutionary Algorithm are two main architecture learning methods used for CNN structure learning \cite{Baker,zhao,Zoph,Barret,zhao,Esteban}. Reinforcement learning uses models' performances on validation dataset as rewards and uses an agent to maximize the cumulative rewards when interacting with an environment. Evolutionary algorithm simulates evolution processes to iteratively improve models' performances.

However, performances of the generated networks are limited by architecture search space which is determined by the algorithm's encoding system. If the encoding system can not represent one architecture, that architecture can never be learned. For example, \cite{Lingxi} used fixed-number stages to represent network structures. Different stages are separated by pooling layer. Each stage is a directed acyclic graph (DAG). In a DAG, nodes represent layers and edges represent operations such as 1$\times$1 convolution and 3$\times$3 convolution. Therefore, the architecture space is directly limited by the fixed-number stages and pooling layers. Because of this, it is impossible to generate many network architectures. For example, it can't generate one network where there is a skip connection between one layer in first stage and another layer in last stage. Some other encoding systems just represent plain architectures which are composed of stacking layers. But plain networks usually suffer from learning problems such as vanishing gradient problem and exploding gradient problem. In this paper, we propose a new encoding system which have little limitations on search space. For example, the order of layers are randomly set. In our encoding system, we use short DNA strands which are composed of a series of DNA molecules (A,G,C,T) to represent layers and long DNA strands composed of short strands to represent architectures.

It is well known that network depth is of crucial importance because it can raise the ``levels'' of features. However, simply stacking layers causes the degradation problem. Skip-connections can solve the degradation problem when deepening networks \cite{srivastava2015highway,srivastava2015training,he2016deep}. ResNet \cite{he2016deep} uses skip-connections via residual blocks. Skip-connections containing in ResNet provide a shorter path between bottom layers and target output layer compared with normal plain architectures. Parameters of bottom layers are thus easier to be trained. DenseNet \cite{huang2017densely} also adopts skip-connections by reusing feature maps. Because skip-connections existing in ResNet and DenseNet ease learning problem and improve model's capacity, we explore neural network architectures with skip-connections.

DNA Computing is a method of computation using molecular biology techniques \cite{braich2000solution,adleman1994molecular,boneh1996computational,kari2000using}. DNA molecules (A, G, C, T) can be used to encode information just like binary strings. DNA strands are composed of DNA molecules and can be regarded as a piece of information. DNA computing uses DNA strands to carry information. It generates large numbers of short DNA strands and then put them into DNA soup. In DNA soup, short DNA strands connect to each other to form longer DNA strands based on base pairing principle if provided suitable reaction environment. After a period of time, the soup contains a sets of candidate DNA strands that represent desired results. Then we can pick DNA strands representing desired results from the soup. For example, ``travelling salesman problem" can be solved by DNA computing \cite{adleman1994molecular}. We can generate different DNA strands and use them to represent a city that have to be visited. Each strand has a linkage with other strands. Within seconds, the strands form bigger ones that represent different travel routes. Then the DNA strands representing longer routes can be eliminated through chemical reaction. The remains are the solutions. In DNA soup, all connecting processes happen at the same time so that DNA computing can reduce reaction time.

In this paper, we use DNA computing algorithm to generate neural network architectures. In DNA computing algorithm, short DNA strand denoted as Layer Strand encodes one layer architecture and a piece of skip-connection information which determines whether the layer has a skip-connection with one of its previous layers. Long strands denoted as Architecture Strands are composed of short Layer Strands via base pairing. Because each layer in network have at most one skip connection with one its previous layers, DNA computing algorithm aims to explore networks with skip-connections. We have little limitations on search space. We don't limit the number of pooling layers and the depth of the architecture. The skip-connections are also randomly set for that any two layers can have a skip-connection. During DNA computing algorithm, we use Layer Strands (representing layers) as our reaction units and learn Architecture Strands (representing architectures) via base paring. After getting models (Architecture Strands) via DNA computing algorithm, we train those models on training data set and select one model according to their performance on the validation data set. We achieve 0.27\% test error on MINST data set and 4.9\% test error on CIFAR-10 data set.

\section{Related Work}
In this section, we introduce convolutional neural networks firstly. Then we introduce reinforcement learning and evolutionary algorithms for structure learning of deep networks.
\subsection{Convolutional Neural Networks}
Convolutional neural networks (CNN) \cite{Alex,Simonyan} have achieved great success in various computer tasks \cite{Backpropagation}. Convolution neural networks are usually composed of convolution layers, pooling layers and fully connected layers. By stacking convolution layers, pooling layers and fully connected layers, we can get plain architectures.

Specialists have tried a lot to improve neural network's capacity and find that increased depth can help a lot. For example, deep models, from depth of sixteen \cite{simonyan2014very} to thirty \cite{ioffe2015batch}, perform well on ImageNet dataset. However, vanishing gradients and exploding gradients prevent models from being deeper \cite{he2016deep}. By normalized initialization \cite{lecun1998efficient,glorot2010understanding,saxe2013exact,he2015delving} and intermediate normalization layers, the problem has been solved a lot and networks can extend to tens of layers. But degradation problem happens when the models become deeper and deeper. Degradation problems mean that models' performance degrades with increased depth.

ResNet  \cite{he2016deep} and DenseNet  \cite{huang2017densely} can solve degradation problems well via skip-connections. Both have gained good results in ImageNet and CIFAR-10. They can also be generalized to many other data sets. Skip-connections make bottom layers have shorter pathes to output layer which makes learning easily and enriches the features. ResNet uses residual blocks to form whole architecture. In each residual block, input layer is added to output layer which is called a skip-connection. So bottom layers in ResNet have a very short path to output layer. The gradients can thus easily and effectively flow to the bottom layers via skip-connections. DenseNet reuses feature maps and increases width of each layer with little increased parameters. The input layer becomes one part of the output layer which can also be called a skip-connection. \\
As the neural networks containing skip-connections perform well in many tasks, we explore neural network architectures with skip-connections. In our encoding system, we don't limit search space. we don't limit the number of pooling layer compared with \cite{Lingxi}. Locations of convolution layers and pooling layers are all randomly set and skip-connections are also randomly set for that one layer can have a skip-connection with any one of its previous layers.
\subsection {Reinforcement Learning and Evolutionary Algorithm}
Even though Resnet and DenseNet perform well in many data sets, network architectures still need to be carefully designed for specific data set. Therefore, how to design a convolutional neural network is a very worthwhile issue. The traditional neural network is designed based on a large number of experimental experience. Recently, more and more researchers focus their research on automatically generating networks and networks have been automatically generated through Reinforcement Learning \cite{Baker} and genetic algorithms \cite{zhao}.

Reinforcement Learning usually uses a meta-controller to determine neural network architectures. It uses architectures' performance on validation data set as reward to update the meta-controller. Thus, a neural network architecture can be treated as a training sample. Because deep learning is a data driving technology, Reinforcement Learning needs a lot of samples to learn a high performance meta-controller. However, training a model spends huge computation resources. Evolutionary Algorithm simulates the process of evolution. It iterates improve its performance by operators such as mutation, crossover and selection. Evolutionary algorithm selects models according to their performances on validation data set. Thus, Evolutionary Algorithm also spends huge computation resources and time.

As the Reinforcement Learning and Evolutionary Algorithm are all data hungry methods and need huge computation sources, we aim to address that we can get a good model via DNA computing algorithm from high quality search space and only train a few of models.
\begin{figure*}[ht]
\centering
\includegraphics[width=3in,height=3in]{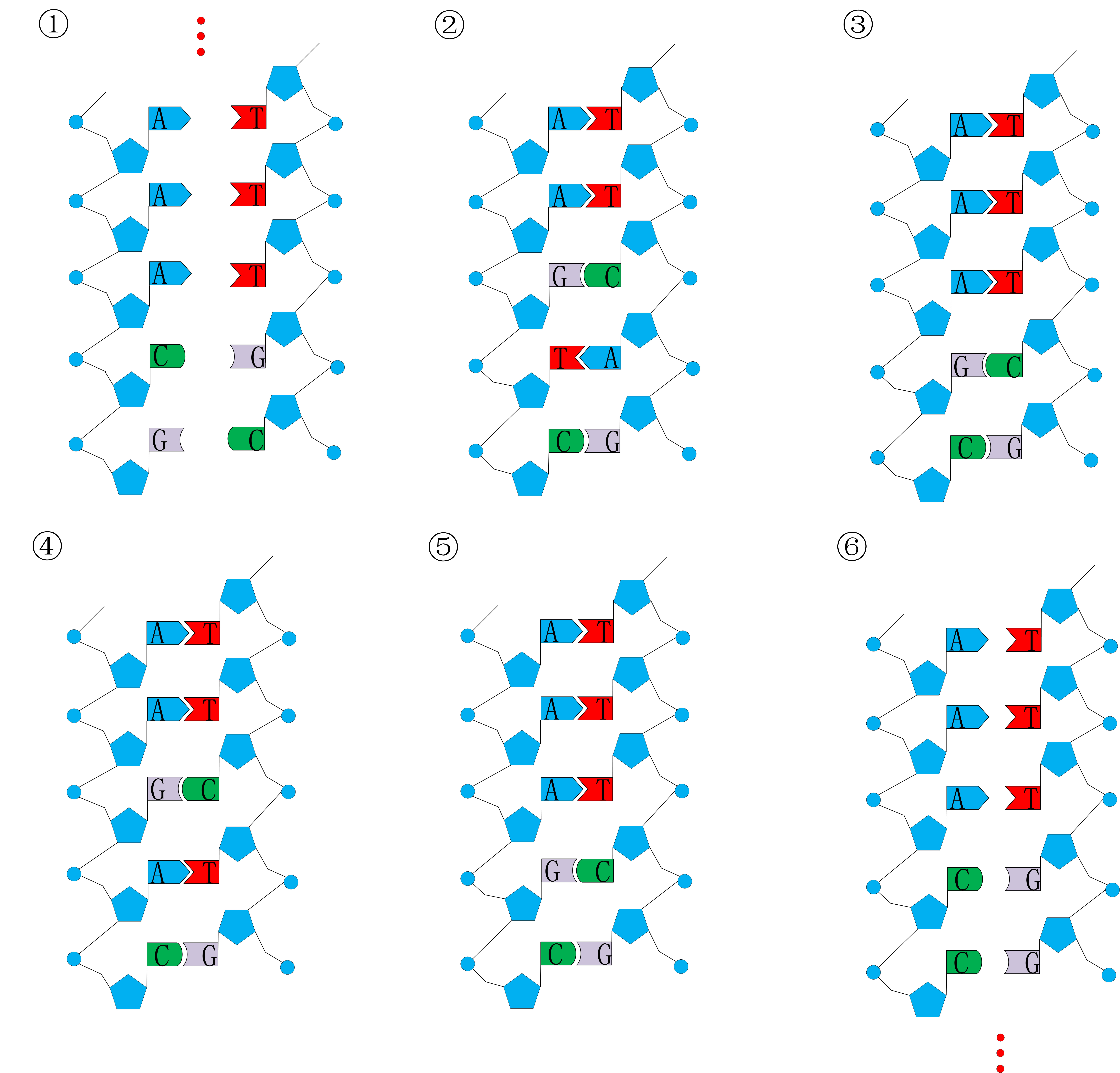}  
\caption{ Layer Strand 9 in one Architecture Strand whose maximum depth is 24 (not including fully connected layers). \textcircled{1}: Left: Head fragment AAACG (9) of Layer Strand 9; Right: Tail of Layer Strand 8. The tail pairs with the head to form a longer strand. \textcircled{2}: Layer type fragment AAGTC (30). Because the result of 30 mod 24 is 6 and 6 is more than 4, the layer belongs to a convolution layer. \textcircled{3}: kernel size fragement AAAGC (6). 6 mod 4 is 2. Because the layer is a convolution layer, thus the number 2 represents 5*5 kernel size. \textcircled{4}: Channel number fragment AAGAC (18). The result of 18 mod 6 is 0, so the channel number is 32. \textcircled{5}: Skip-connection fragment AAAGC (6). 6 mod 9 is 6. So there is a skip-connection between layer 6 and layer 9. \textcircled{6}: Left: The head fragment of layer 10. Right: The tail fragment of layer 9.}
\label{f1}
\end{figure*}
\section{Our Approach}
In this section, we introduce how DNA strands encode neural network architectures. And then, we introduce how to generate architectures using DNA Computing Algorithm. After introduction of model generation, we introduce how to train the learned models.
\subsection{Coding System}\label{codesys}
DNA computing algorithm uses DNA strands to encode neural network architectures and use DNA computing algorithm to generate the strands that represent architectures. In our DNA computing algorithm, we use short DNA strands denoted as Layer Strands to represent layer architectures (convolution layers and pooling layers) and long strands denoted as Architecture Strands to represent overall neural network architectures. In DNA computing algorithm, Layer Strands can form Architecture Strands via base paring between the exposed heads and tails of different Layer Strands, which is like the process that stacking layers to form architectures. The skip-connection information is encoded in each Layer Strand. And one layer can have at most one skip-connection with one of its previous layer. Those architectures with skip-connections form architecture space learned by DNA computing algorithm. We must emphasize we don't limit the search space for that we only set the maximum depth of the model. The number of pooling layers, location of convolution layers and skip-connection in each layer are all random. The detail of encoding method are described below.

Because the similarity between pooling layer and convolution layer, pooling layer and convolution layer are represented by Layer Strands that have same constructions. Each Layer Strand is composed of 6 DNA-fragments representing specific parameters of pooling layer or convolution layer, such as kernel size or channel number. Those 6 DNA-fragments include head fragment, layer type fragment, kernel size fragment, channel number fragment, skip-connection fragment and tail fragment. The head fragment represents layer number. For example, some Layer Strands represent first layer while some Layer strands represent second layer and so on. If one Layer Strand represent ith layer, it is called Layer Strand i-1. The tail fragment are specially designed to pair the head fragment of its previous Layer strand so that two Layer Strands can form a longer strand by base paring method. The layer type fragment determines which kinds of layer this strand represents, convolution layer or pooling layer. The channel number fragment determines the number of channels. The number of channels is chosen from \{32, 64, 96, 128, 160, 192\}. The channel number fragment of pooling layer is useless for that the channel number of pooling layer is same as its previous layer. The kernel size fragment determines the kernel size. For convolution layer, the kernel size is chosen from \{1$\times$1, {3$\times$3}, {5$\times$5}, {7$\times7$}\}. For pooling layer, the kernel size is chosen from \{2$\times$2, 3$\times$3\}. As for skip-connection fragment, it determines whether this layer has a skip-connection with one of its previous layer. Each layer has at most one skip-connection. Six DNA fragments are arranged according to head, layer type, kernel size, channel number, skip-connection, and tail order. In reality, DNA is a double-stranded structure with exposed head and tail. So in addition to the head and tail fragments, the other four fragments belong to double-stranded structures. The head and tail fragments are exposed with single layer structure. Thus different Layer Strands can be connected between their exposed head and tail fragments' base paring. Six fragments compose a Layer strand which is a basic unit in DNA computing.

Just as Figure \ref{f1} shows, each fragment of Layer Strand is composed of five pairs of molecules or five single molecules (head and tail). For double-stranded structure fragments, only the strand along the same side as the tail and head is useful during decoding. Therefore, the length of a Layer Strand is 30. We use molecules A, G, C, T to represent 0, 1, 2, 3 respectively. According to the quaternary decoding method, five molecules represent integers from 0 to 1023. Thus one fragment can encode 1024 (4$^5$) kinds of information. There are redundant integers in each strands and we use different groups of integers to represent different parameter values in the four double-structured fragments so that all the permutation of five molecules can be utilized. The detail will be described below.

Hyper-parameter N specifies the maximum number of layers in each neural network. Each fragment can be translated into a real number and the specific parameter are determined by the real number. The head fragment can be translated into integer n and the integer n represents layer n. In generation, only Layer Strands represent layer 0 (first layer) to layer N-1 can be generated. Thus, only head fragments represent layer 0 to layer N-1 can be generated. The Layer Strand representing layer i can be denoted as Layer Strand i. In this way, we can define maximum depth of neural network. Because the tail fragment representing layer L needs to pair the head fragment of layer L+1 to form a longer strand, the tail fragment of layer L is thus determined by the tail fragment of layer L+1. For example, the head fragment representing layer one is AAAAG (can be translated into number 1), then the tail fragment of the layer 0 must be TTTTC (A pairs with T and G pairs with C). Thus, the two Layer Strands can be connected by base paring between exposed AAAAG and TTTTC single-layer fragments to form a long strand. So, only N kinds of tail fragments corresponding to head tails can be generated. As for the other four kinds of fragment, they are not limited in generation. All kinds of permutation of five molecules can be generated.

During decoding, one fragment is translated into real number n and the concrete meanings of fragments are determined by integer groups which the integers belong to. Layer type fragment can be translated into n$_t$. If the result of n$_t$ mod N is less than 4, the layer belong to pooling layer. Otherwise, it is a convolution layer. That means we may get about 4 pooling layers in the N layers. The kernel size fragment can be translated into number n$_k$. If the layer is a convolution layer, the result of n$_k$ mod 4 determines the kernel size and 0, 1, 2, 3 represent 1$\times$1, {3$\times$3}, {5$\times$5}, {7$\times7$} kernel sizes respectively. If the layer is a pooling layer, the result of n$_k$ mod 2 determines the kernel size and 0, 1 represent 2$\times$2, 3$\times$3 kernel sizes respectively. The channel number fragment can be translated into real number n$_c$. The results of n$_c$ mod 6 determine the channel numbers and 0, 1, 2, 3, 4, 5 represent 32, 64, 96, 128, 160, 192 respectively. The skip-connection fragment can be translated into real number n$_s$. If the layer number is L and the result of n$_c$ mod L is l, there is a skip-connection between the layer L and layer l. l should be less than L-1. Otherwise, the skip-connection fragments are useless.
\begin{figure}[h]
\centering
\includegraphics[width=3in,height=5.8in]{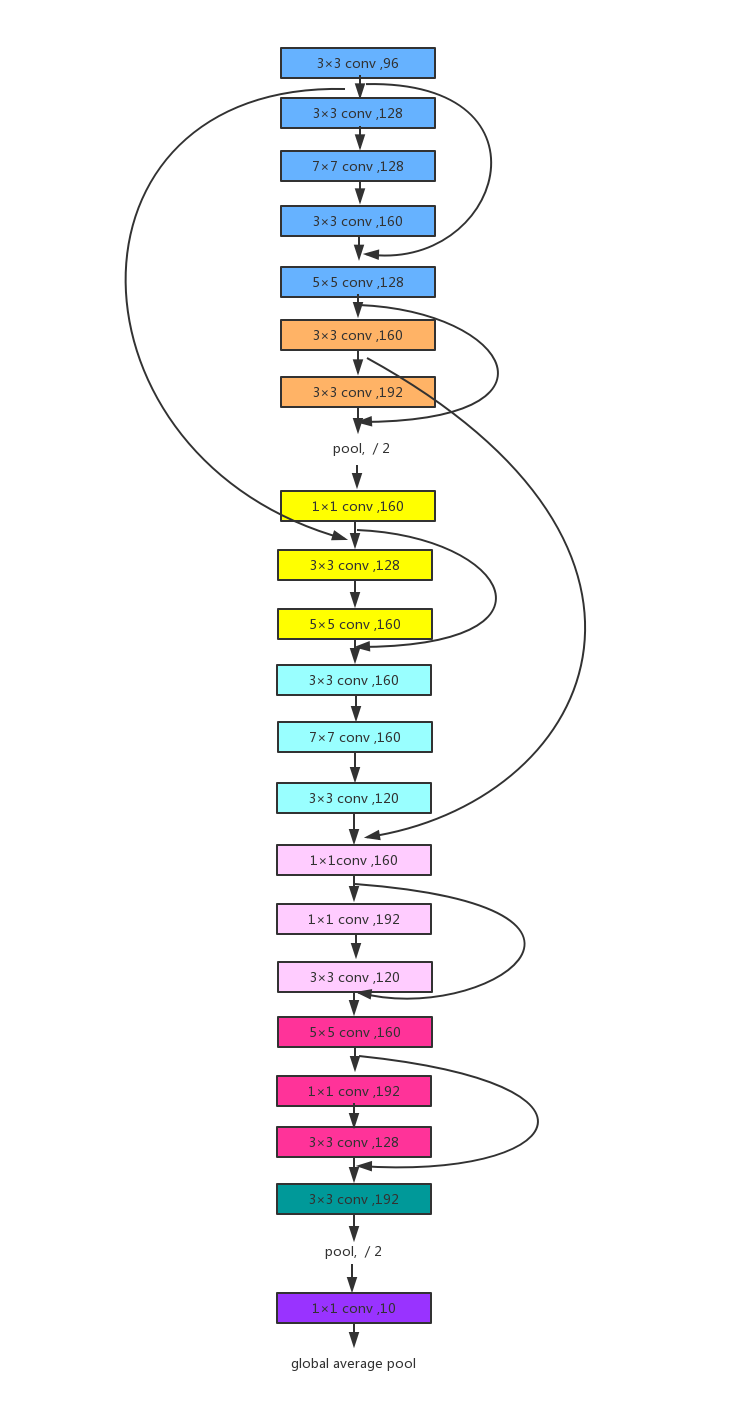}  
\caption{One of our models on CIFAR-10 that gains high accuracy. }
\label{f3}
\end{figure}
\subsection{Generation via DNA Computing Algorithm}
Layer Strands are our basic reaction units of DNA computing algorithm. Hyper-parameter P defines the number of architectures need to be generated. In generation, we generate P Layer Strands i (0$\leq$i$\le$N-1). We can get P$\times$N Layer Strands. In each Layer Strand, only head fragments and tail fragments are generated specially while other fragments are randomly generated. 

We then put all the Layer Strands into DNA soup. In DNA soup, Layer Strand i and Layer Strand i+1 are connected via base paring between their exposed head and tail. The head fragment of Layer Strand i pairs with the tail of Layer Strand i+1 to form a double-structure fragment. Thus, the two strands form a longer strand. All the connection processes can happen at the same time.

The process finish within seconds. Then we can select DNA strands from DNA soup and eliminate the strands whose start part is not Layer Strand 0. After we get Architecture Strands, we translate them into real architectures. In the translating, we add a fully connected layer at the end of each architecture. We train them on training data set and select one model based on their performances on validation data set. We simulate DNA computing algorithm via computer. Algorithm \ref{alg:1} illustrates the process of DNA computing algorithm.
\begin{algorithm}[ht]
	\renewcommand{\algorithmicrequire}{\textbf{Input:}}
	\renewcommand{\algorithmicensure}{\textbf{Output:}}
	\caption{DNA Computing Algorithm}
	\label{alg:1}
	\begin{algorithmic}[1]
		\REQUIRE {Dataset, maximum number of layers in one architecture (N), number of Layer Strands representing one layer (P);\\}
		\ENSURE   The network structure with the highest accuracy on test data set.
        \FOR { i=1 to N }
        \FOR { j=1 to P }
		 \STATE {Generate one Layer Strand i randomly. }
        \ENDFOR
        \ENDFOR
		\STATE{Put all the Layer Strands into DNA soup and provide proper reaction environment.}\\
        \STATE{Select Architecture Strands from DNA soup. Count the number of Architecture Strands and get number Num. }
		\FOR{ i=1 to Number}
		\STATE {{S}=Generate real CNN network.}
        \ENDFOR
        \STATE {{G} = Randomly select 100 models.}
		\FOR { i=1 to 100 }
		 \STATE { Training network i in {G} on training data set and record its network structure and final accuracy on validation data set}
        \ENDFOR
		\STATE {Select the model R from G that has highest validation accuracy.}
        \STATE {Train model R on the whole training data set and validation data set and get its accuracy A on test data set.}
		\STATE \textbf{return} R, A
	\end{algorithmic}
\end{algorithm}

\subsection{Model Training}
Just as algorithm \ref{alg:1}, we train all the models on the training data set and get their accuracies on validation data set after getting model via DNA Computing Algorithm. We select the best model according to their performances on validation data set. After getting the best model, we merge the training data set and validation data set and train the model again. we then use the model's performance on test data set as our algorithm output. The training details are described below.

To compress the search space, we used a pre-activated convolution unit (PCC). That's to say, we use batch normalization(BN) \cite{ioffe2015batch} and ReLU activation \cite{Alex} before convolution operation. The stride of convolution layer is set as 1 while the stride of pooling layer is set as 2. As for skip-connections, if the layer i and layer j (i$\leq$j) has a skip-connection. We then add layer i and layer j as output of layer j. If the two layers' channels don't map, we use 1$\times$1 convolution with stride 1 to change channels. If the feature map size don't map, we use 1$\times$1 convolution with stride 2 to down sample.

As for optimizer, we use momentum optimizer with momentum set to 0.9. The initial learning rate is 0.1 and the weight decay is 0.0001. The total training epochs is 60. In the tenth epoch, the learning rates is set to 0.01. In the thirtieth epoch, the learning rates is set to 0.001. 

The models are trained for at most 60 epochs on training sets (CIFAR-10). As for MNIST, 10 epochs is enough for the models to be converged. Carefully designing total epochs reduces huge time. At training time, we find that with a fixed learning rate, the algorithm is converged around some epochs and had a little progress on the further epochs. If we do not decrease the learning rate, the accuracy increase little. So it's necessary to immediately decrease the learning rate after the accuracy increase slowly. If the learning rate is set properly, it reduce huge time. During the training process, we used L2 regularization.

\begin{figure}[h]
\centering
\includegraphics[width=3in]{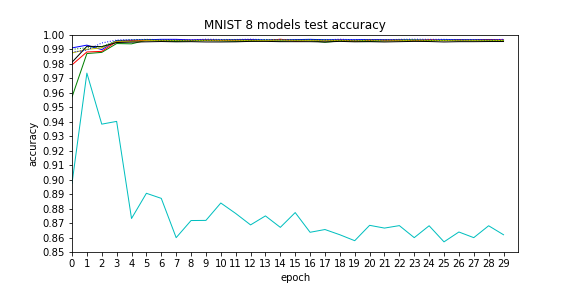}  
\caption{The learning curve of MNIST and CIFAR10 is very similar and proves the early stop strategy can also be used on MNIST.}
\label{f6}
\end{figure}

\begin{figure}[h]
\centering
\includegraphics[width=3in]{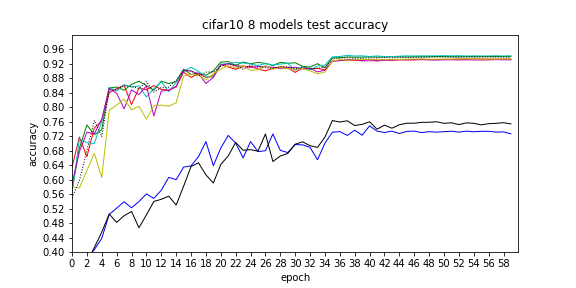}  
\caption{  We randomly chose eight of all our generated models without early stop strategy. As the figure shows, the competitive models perform well all the time, but the poor models are contrary. That demonstrates we can eliminate the poor model according their learning curve. }
\label{f5}
\end{figure}
\subsection{Early Stop Strategy}
In order to further reduce running time, we used early stop strategy. Just as shown in Figure \ref{f5} and Figure \ref{f6}, the epochs can be reduced by carefully design. We have two main discoveries. Firstly, all the models have similar learning curves. That indicates us we can eliminate models without finishing all the training epochs. Secondly, most of models in our search space perform similarly showing that we can get a high performance model with fewer times of training. We can stop training if the model can not perform well until the specified epoch. The early stop strategy eliminates poor performance models with high probability for that good models usually perform well on an early stage. This strategy reduces the huge time and has little impact on accuracy. We set three model performance thresholds on CIFAR-10 data set that are trained for at most 60 epochs. (1) 10th epoch, 80\% test accuracy. (2) 20th epoch, 85\% accuracy. (3) 45th epoch, 90\% accuracy. If the model does not reach the specified accuracy after the there epochs respectively, the models will be eliminated.

\section{Results}
In this section, we introduce our results on CIFAR-10 and MNIST data sets. We simulate DNA computing algorithm by computer.
\begin{table}[h]
  \centering
  \setlength{\tabcolsep}{0.8mm}
    \scalebox{0.9}[0.9]{
  \begin{tabular} {|l|l|l|l|}
  \hline
  Model&MNIST\\
  \hline
  Lecun\emph{et.al}\cite{lecun1998gradient}&0.7\\
  \hline
  Lauer\emph{et.al}\cite{lauer2007trainable}&0.54\\
  \hline
  Jarrett\emph{et.al}\cite{jarrett2009best}&0.53\\
  \hline
  Ranzato\emph{et.al}\cite{poultney2007efficient}&0.39\\
  \hline
  Cirecsan\emph{et.al}\cite{cirecsan2012multi}&0.23\\

  \hline
  {Our Method}&{0.27}\\
  \hline
  \end{tabular}
  }
  \caption{Comparison of the recognition error rate (\%) on MNIST.}
\end{table}

\subsection{Results on the MNIST Data Set}
The MNIST database of handwritten digits has a training set of 60,000 examples, and a test set of 10,000 examples. We test our algorithm on MNIST data set in order to reduce time. We divide the training data set into two parts. 55,000 images are used as training set while 5,000 images are used as validation data set. After we get the best model according their performances on validation data set, the whole 60,000 images are used for training the model. We then test the model on test data set and use the accuracy as our method's final output.

\begin{figure}[h]
\centering
\includegraphics[width=3in]{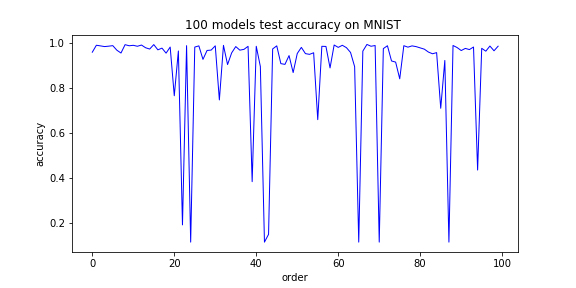}  
\caption{One hundred models test accuracy on MNIST. Most of models have gained similar performances.}
\label{f7}
\end{figure}

We use DNA Computing algorithm to generate model architectures. Many of our models have gained test accuracies higher than 99.60\% and the highest test accuracy is 99.73\%. We show one hundred models' learning in Figure \ref{f7}. Only few of them perform poorly indicating that our neural network architectures with random skip-connection composes a high performance search space.
\begin{table}[h]
  \centering
  \setlength{\tabcolsep}{1.0mm}
   \scalebox{0.9}[0.9]{
  \begin{tabular}{p{1.2cm}|p{6cm}|p{1.4cm}}
  \hline
  &Model&CIFAR-10\\
  \hline
  &Maxout \cite{goodfellow2013maxout}&9.38\\
  human&ResNet(depth=110) \cite{he2016deep}&6.61\\
  designed&ResNet(pre-activation)\cite{he2016identity}&4.62\\
  &DenseNet(L=40,k=12)\cite{huang2017densely}&5.24\\

  \hline
  &GeNet\#2(G-50) \cite{Lingxi}&7.10\\
  auto&Large-Scale Evolution\cite{Esteban}&5.40\\
  designed&NAS (depth=39)\cite{Lee}&6.05\\
  &MetaQNN\cite{Baker}&6.92\\
  &EAS\cite{cai2018efficient}&4.23\\
  \hline
  &{Our Method(depth=49)}& {4.9}\\
  \hline
  \end{tabular}
  }
  \caption{Comparison of the recognition error rate (\%) on CIFAR-10.}

\end{table}

\subsection{Results on the CIFAR-10 Data Set}
The CIFAR-10 dataset consists of 60000 32x32 colour images in 10 classes, with 6000 images per class. There are 50,000 training images and 10,000 test images. The training data set are divided into two parts, new training data set(55,000 images) and validation data set(5,000 images).

\begin{table}[h]
  \centering
  \setlength{\tabcolsep}{0.8mm}
    \scalebox{0.9}[0.9]{
  \begin{tabular} {p{5.5cm}|p{2cm}}
  \hline
  Model&CIFAR-10\\
  \hline
  DNA computing algorithm (depth=13)&7.34\\
  \hline
  DNA computing algorithm (depth=17)&6.7\\
  \hline
  DNA computing algorithm (depth=21)&6.1\\
  \hline
  DNA computing algorithm (depth=25)&5.65\\
  \hline
  DNA computing algorithm (depth=49)&4.9\\
  \hline
  \end{tabular}
  }
  \caption{Comparison of our models' recognition error rate (\%) with different depth of models on CIFAR-10 data set.}
\end{table}
Our best model gain 95.10\% test accuracy with 49 layers architectures composed of convolution layers, pooling layers and a fully connected layer. Note that we use data augmentation (flip, crop, and data normalization). With few fully trained models, we still gain high test accuracy. It proves that learning models from carefully designed search space via DNA computing algorithm can gain high accuracy. It is possible for non-experts without expertise to get a high accuracy models in specific task. One of our high performance models are showed in Figure \ref{f3}.

\section{Conclusion}
We propose DNA computing algorithm to learn neural networks from a well defined architecture search space. Our search space is defined by architectures with skip-connections. During training models, we use early stop strategy which saves time and computation sources. We find most models perform similarly in our search space and have similar learning curves. We prove that learning neural networks via DNA computing algorithm is feasible and gain high accuracy. And we find that local minimal is not of importance during training models and using early stop strategy can eliminate models just after several epochs of training. We conduct the algorithm in two data sets (CIFAR-10 and MNIST) and we get competitive results in comparison with evolutionary algorithm and Reinforcement Learning but training fewer models.
We simulate DNA computing algorithm by computer. In future work, We consider doing biochemical experiments to verify the feasibility of the method. 
\bibliographystyle{aaai}
\bibliography{mybib}

\end{document}